\documentclass{article} 
\usepackage{colm2024_conference}
\usepackage{graphicx}
\usepackage{microtype}
\usepackage{hyperref}
\usepackage{url}
\usepackage{booktabs}
\usepackage{amsmath}

\title{On-Device Language Models: A Comprehensive Review}

\author{Jiajun Xu \thanks{Equal contribution} \\
Meta\\
\texttt{\{jjxu217\}@meta.com} \\
\And
Zhiyuan Li$^*$\\
Nexa AI\\
\texttt{\{zack\}@nexa4ai.com}
\And
Wei Chen$^*$ \\
Nexa AI\\
\texttt{\{alexchen\}@nexa4ai.com}
\And
Qun Wang$^*$ \\
Computer Science Department, San Francisco State University\\
\texttt{\{qunwang\}@sfsu}
\And
Xin Gao$^*$ \\
University of North Texas\\
\texttt{\{xingao\}@my.unt.edu}
\And
Qi Cai$^*$ \\
University of North Texas\\
\texttt{\{qicai\}@my.unt.edu}
\And
Ziyuan Ling$^*$ \\
Nexa AI\\
\texttt{\{rita\}@nexa4ai.com}
}

\colmfinalcopy 
\begin{document}

\maketitle
\begin{abstract}
The advent of large language models (LLMs) revolutionized natural language processing applications, and running LLMs on edge devices has become increasingly attractive for reasons including reduced latency, data localization, and personalized user experiences. This comprehensive review examines the challenges of deploying computationally expensive LLMs on resource-constrained devices and explores innovative solutions across multiple domains. The paper investigates the development of on-device language models, their efficient architectures, including parameter sharing and modular designs, as well as state-of-the-art compression techniques like quantization, pruning, and knowledge distillation. Hardware acceleration strategies and collaborative edge-cloud deployment approaches are analyzed, highlighting the intricate balance between performance and resource utilization. Case studies of on-device language models from major mobile manufacturers demonstrate real-world applications and potential benefits. The review also addresses critical aspects such as adaptive learning, multi-modal capabilities, and personalization. By identifying key research directions and open challenges, this paper provides a roadmap for future advancements in on-device language models, emphasizing the need for interdisciplinary efforts to realize the full potential of ubiquitous, intelligent computing while ensuring responsible and ethical deployment. For a comprehensive review of research work and educational resources on on-device large language models (LLMs), please visit \href{https://github.com/NexaAI/Awesome-LLMs-on-device}{https://github.com/NexaAI/Awesome-LLMs-on-device}. To download and run on-device LLMs, visit \href{https://www.nexaai.com/models}{https://www.nexaai.com/models}.

\end{abstract}

\section{Introduction}

The emergence of Large Language Models (LLMs) has catalyzed a transformative shift in natural language processing (NLP) applications. By leveraging the transformer architecture \citep{vaswani2017attention}, LLMs such as OpenAI's GPT series \citep{radford2019language, brown2020language, GPT4} and Meta's LLaMA series \citep{llama1, llama2, llama3, llama3-1} have demonstrated unparalleled proficiency in understanding and generating human-like text, profoundly influencing fields ranging from automated customer support to advanced content creation. The ability of these models to seamlessly perform a variety of NLP tasks has positioned them as the backbone of modern AI-driven applications \citep{wu2023autogen, openagi, nam2024using, zheng2024judging, yang2024harnessing}.

However, the traditional deployment of LLMs predominantly on cloud servers presents several challenges, particularly in terms of latency, security, and the need for continuous Internet connectivity. These concerns are driving the burgeoning interest in deploying LLMs on edge devices—a shift that promises reduced response times, and personalized user experiences directly on user devices such as smartphones, automotive systems, and personal wearables. This paradigm shift not only aligns with the increasing user demand for immediate and personalized assistance but also mitigates the bandwidth and energy costs associated with cloud computing.

\begin{figure}[h]
    \centering
    \includegraphics[width=\textwidth]{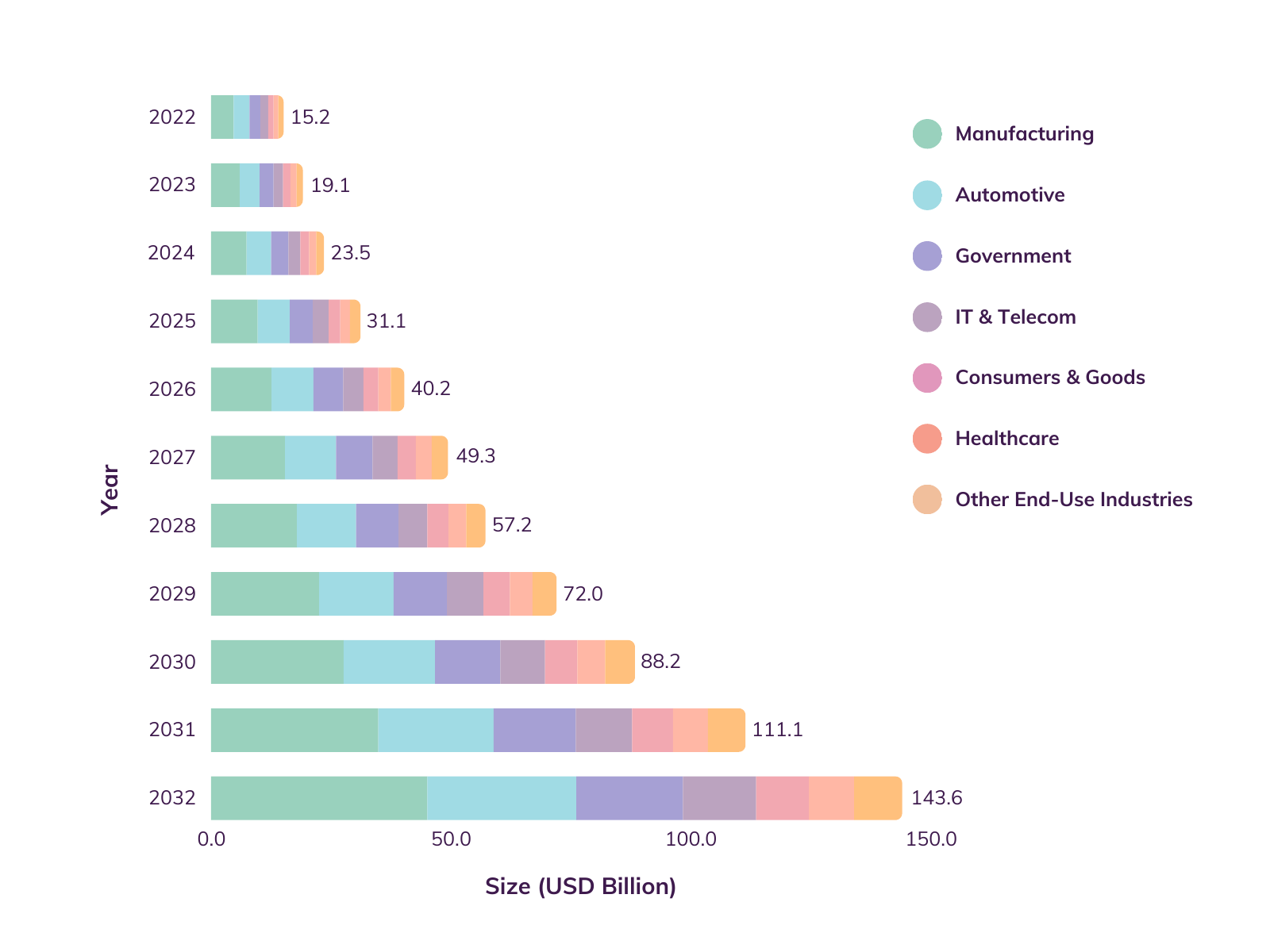}
    \caption{The global market size for on-device edge AI, by end-user, from 2022 to 2032, in USD Billion. The market will grow at the CAGR of 25.9\%. The forecasted market size for 2032 is \$143.6B \citep{Marketus2024}.}
    \label{market_size}
\end{figure}
The growing interest in on-device AI deployment is reflected in the rapidly expanding edge AI market. As illustrated in Figure \ref{market_size}, the edge AI market is projected to experience substantial growth across various sectors from 2022 to 2032. The market size is expected to increase from \$15.2 billion in 2022 to \$143.6 billion by 2032, representing a nearly tenfold growth over a decade \citep{Marketus2024}. This growth spans multiple industries, with manufacturing, automotive, and government sectors showing significant contributions. The projected market expansion underscores the increasing demand for edge AI solutions, including on-device language models, driven by the need for faster, more private, and efficient AI capabilities across diverse applications. This market trend aligns with the technological push towards more localized AI processing, further emphasizing the importance of developing efficient on-device LLM solutions.

Despite the compelling advantages, integrating computationally intensive language models within the constraints of edge devices poses significant challenges. The primary obstacles include limited computational power, reduced memory capacity, and energy constraints, which collectively complicate the direct adoption of cloud-based LLM architectures. For instance, executing a state-of-the-art 405-billion parameters model \citep{llama3-1} on a smartphone would be unfeasible without substantial compromises in model performance and energy efficiency.

\begin{figure}[!htb]
    \centering
\includegraphics[width=0.99\textwidth,height=0.89\textheight]{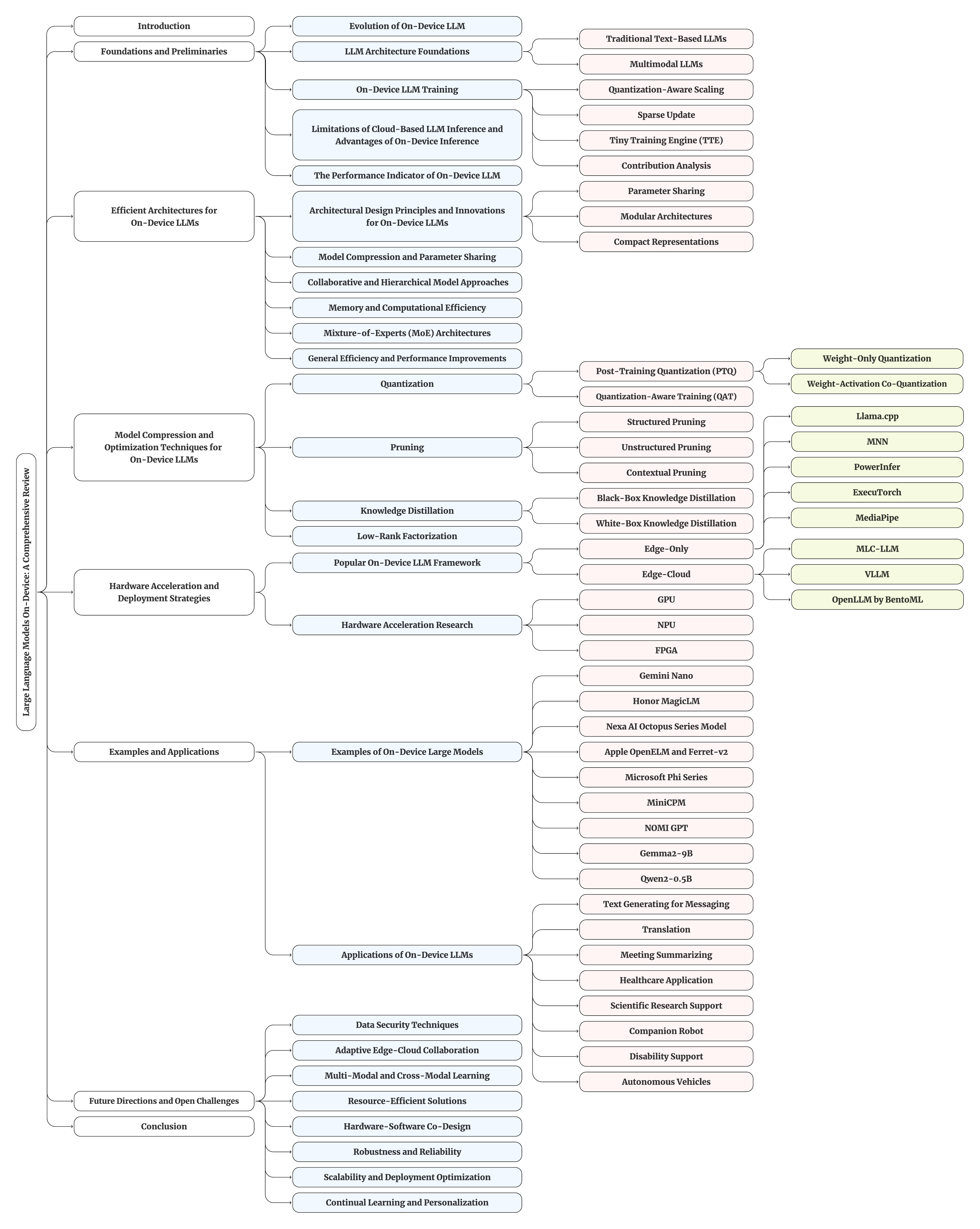}
    \caption{The architecture of this paper}
    \label{outline}
\end{figure}

This review paper provides a comprehensive exploration of the current strategies and advancements in the deployment of LLMs on edge devices. We aim to critically analyze the various techniques and architectures that have been developed to adapt LLMs to the constraints of edge computing. This includes a detailed examination of model compression techniques, energy-efficient computing strategies, and the development of novel lightweight model architectures. Furthermore, the paper will delve into deployment strategies that enable the effective use of LLMs in edge scenarios, highlighting key industry applications and the resulting benefits.

Through this review, we intend to illuminate the pathways and challenges in transitioning from cloud-based to on-device language models, providing insights into how this shift could redefine the landscape of applications and AI accessibility. The structure of this paper is illustrated in Fig. \ref{outline}. We begin by exploring the foundations and preliminaries in Section \ref{sec: pre}, including the evolution of LLMs on-device, architectural foundations, and on-device training techniques. Section \ref{sec: arch} delves into efficient architectures for on-device language models, discussing innovative design principles, model compression, and collaborative approaches. Section \ref{sec: comp} continues with an in-depth examination of model compression and optimization techniques, covering quantization, pruning, knowledge distillation, and low-rank factorization. Section \ref{sec: hardware} investigates hardware acceleration and deployment strategies, highlighting popular on-device LLM frameworks and hardware-specific optimizations. To contextualize these advancements, in Section \ref{sec: eg}, we present examples of existing on-device language models and their real-world applications across various domains. Finally, Section \ref{sec: directions} discusses future directions and open challenges in the field, and Section \ref{sec: con} concludes our review. By focusing on the intersection of LLM capabilities and edge computing requirements, this paper contributes to the ongoing discourse in AI research, offering a comprehensive perspective on achieving the delicate balance between model performance and computational efficiency in resource-constrained environments.

\section{Foundations and Preliminaries}
\label{sec: pre}
\subsection{Evolution of On-Device LLMs}

\begin{figure}[h]
    \centering
    \includegraphics[width=\textwidth]{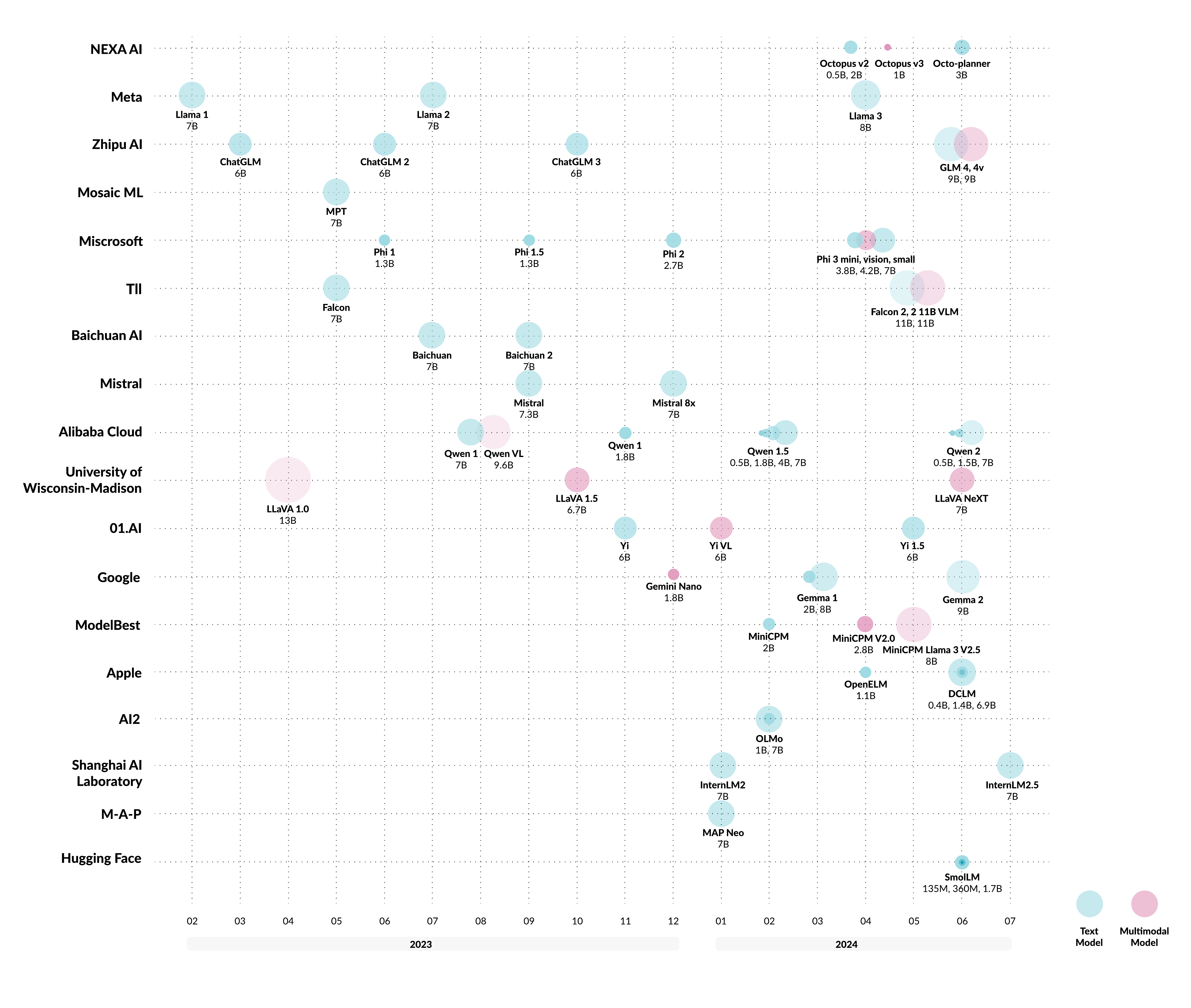}
    \caption{Summary of on-device LLMs' evolution}
    \label{evolution}
\end{figure}

The evolution of on-device LLMs is a process closely linked to technological progress. 
Figure \ref{evolution} provides a comprehensive timeline of on-device language model development since 2023, illustrating the rapid advancement in this field.
As shown in the figure, the exploration and experimentation of large language models on the edge began in earnest in 2023. We saw the emergence of several influential model series with parameters below 10B, making it possible for LLMs to run on edge devices. Notable examples include:

\begin{itemize}
    \item Meta's LLaMA series (\cite{llama1,llama2,llama3, llama3-1})
    \item Microsoft's Phi series (\cite{gunasekar2023textbooks,li2023textbooks,abdin2024phi3})
    \item Zhipu AI's ChatGLM series (\cite{chatglm})
    \item Alibaba's Qwen series (\cite{bai2023qwen,Qwen2})
    \item 01.AI's Yi series (\cite{young2024yi,Yi1.5})
    \item Mistral's series (\cite{jiang2023mistral,jiang2024mixtral})
    \item Shanghai AI Laboratory's InternLM series (\cite{team2023internlm, internlm2})
\end{itemize}

In addition, there are also models such as Falcon released by TII \citep{almazrouei2023falcon} and the MPT model released by Mosaic ML \citep{MPT} that have participated in the competition of such models. Although the performance of these small-parameter models is not as good as that of traditional large-parameter models, they make it possible for LLMs to run on edge devices. Their appearance marks the importance of the language model industry to the application scenarios of edge devices using LLMs. At the same time, with the application of technologies such as mixed experts, quantization, and compression, the performance of small-parameter models is constantly making great progress while maintaining the parameter volume.

Figure \ref{evolution} also highlights the emergence of multimodal models since 2023, such as the LLaVa series \citep{liu2024improved, liu2024visual}, QwenVL \citep{Qwen-vl}, Gemini Nano \citep{team2023gemini}, and Yi VL \citep{young2024yi}. These models represent valuable attempts to deploy multimodal LLMs on the edge, adapting to more complex and changing user scenarios on mobile devices.

Entering 2024, the pace of innovation accelerated, as evident from the dense cluster of new models in the figure's rightmost section. This period saw the introduction of:
\begin{itemize}
    \item Nexa AI's Octopus series \citep{chen2024octopus2,chen2024octopus3, chen2024octopus4}
    \item ModelBest's MiniCPM series \citep{hu2024minicpm, MiniCPM}
    \item Google's Gemma series \citep{gemma, Gemma2}
    \item Apple's OpenELM \citep{mehtaOpenELMEfficientLanguage2024} and DataComp-LM \citep{dclm}
    \item AI2's OLMo \citep{dolma, olmo}
\end{itemize}

Figure \ref{evolution} clearly shows an increased focus on multimodal capabilities in 2024, with many new models offering both text and multimodal functionalities to address diverse task-processing scenarios. As illustrated by the variety and progression of models, on-device language models are rapidly evolving and diversifying. This trend, coupled with the continuous maturation of intelligent hardware and software technologies, enables the integration of these models into smartphones, Internet-connected cars, computers, robots, and other terminal equipment, showcasing their growing application potential and value.

\subsection{LLM Architecture Foundations}

\begin{enumerate}
    \item \textbf{Traditional text-based LLMs:}
Let's start where it all began. Transformer is a deep learning model based on an attention mechanism \citep{vaswani2017attention}, widely used to process sequential data, especially in natural language processing tasks. It consists of two parts: an encoder and a decoder. Nowadays, popular large language models mainly use decoder-only architecture \citep{fu2023decoder}, representing models such as GPT (Generative Pre-trained Transformer), LLaMA (Large Language Model Meta AI), etc. The GPT model consists of multiple decoder layers \citep{radford2018improving, radford2019language, brown2020language}, and each decoder layer consists of a self-attention mechanism. The GPT model also applies layer normalization after each sub-layer \citep{floridi2020gpt}. In contrast, LLaMA applies normalization \citep{ioffe2015batch, zhang2019root, xiong2020layer} before each sub-layer operation, which helps to improve the stability of the training process \citep{llama1}.
In terms of the application of attention mechanisms, the GPT model uses the standard self-attention mechanism, which allows the model to consider information from all positions in the input sequence when generating the sequence, while LLaMA uses Group Query Attention (GQA) \citep{ainslie2023gqa}, which is an optimization technique that reduces the computational and memory footprint of the model and improves efficiency.

The concept MoE (Mixture of Expert), originated in 1991 \citep{jacobs1991adaptive}, plays a key role in today's language models pre-training. It enables efficient pre-training with far less computational resources than are required for dense models. The mechanism consists of two key components: a sparse MoE layer containing a number of ``experts", each of which is a separate neural network in its own right \citep{shazeer2017outrageously, chen2022towards, du2022glam}; and a gating network or routing: this component is used to determine which tokens are sent to which expert model for processing. Architecture replaces each feed-forward network (FFN) layer in a traditional Transformer model with a MoE layer, which consists of two core components: a gating network and a number of experts \citep{MoE}.

    \item \textbf{Multimodal LLMs:}
With the powerful learning architecture of Transformer, large multimodal models can process multiple different modalities at the same time, such as text, images, sounds, data tables, etc \citep{xie2024large, wu2023multimodal}. Its internal operating mechanisms are as follows: 

A) Use standard cross-attention layers to perform deep fusion of multimodal inputs in the internal layers of the model (such as MultiModal-GPT \citep{gong2023multimodal})

B) Use custom-designed layers to perform deep fusion of multimodal inputs in the internal layers of the model (LLaMA-Adapter (\cite{zhang2023llama}), MoE-LLaVa (\cite{Moe-llava}))

C) Perform early fusion of multimodal inputs at the input stage of the model, using modality-specific encoders (LLaVa \citep{liu2024visual}, Qwen-VL \citep{bai2023qwen})

D) Perform early fusion at the input stage, but use tokenization techniques (such as tokenizers) to handle modalities \citep{wadekar2024evolution}.
\end{enumerate}

\subsection{On-Device LLMs Training}
Deploying large language models (LLMs) on resource-constrained devices poses challenges such as limited memory and computational power (\cite{loukas2023making}). To address these issues, collaborative and hierarchical model approaches offer innovative solutions by distributing computational load and utilizing models with varying capabilities.

Classic methods for training on resource-constrained devices include:
\begin{enumerate}
    \item{Quantization-aware scaling:} Stabilize the training process by automatically scaling the gradients of tensors with different bit precisions, solve the problem of inconsistent gradient scales of tensors with different bit widths in the quantization graph, and make the training accuracy of the quantized model comparable to that of the floating-point model \citep{nagel2022overcoming, huang2024billm}. 

    \item{Sparse update:} Selectively update the weights of a portion of the layers in the network, skip the gradient calculations of less important layers and sub-tensors, thereby reducing memory usage and computational costs \citep{liu2023deja, ansell2024scaling}.

    \item{Tiny Training Engine (TTE):} Includes redundant nodes in the reverse graph, such as gradient nodes that freeze weights, and reorder operations to achieve in-place updates \citep{lin2023tiny, khouas2024training}.

\item{Contribution analysis:} Automatically determine the sparse update scheme, that is, determine which parameters (weights/biases) contribute the most to downstream accuracy, so as to select which layers or parts of tensors should be updated under a limited memory budget \citep{lin2022device, ren2024timechat, zeng2023agenttuning}.
\end{enumerate}

\subsection{Limitations of Cloud-Based LLM Inference and Advantages of On-Device Inference}
\begin{figure}[h]
    \centering
    \includegraphics[width=0.8\textwidth]{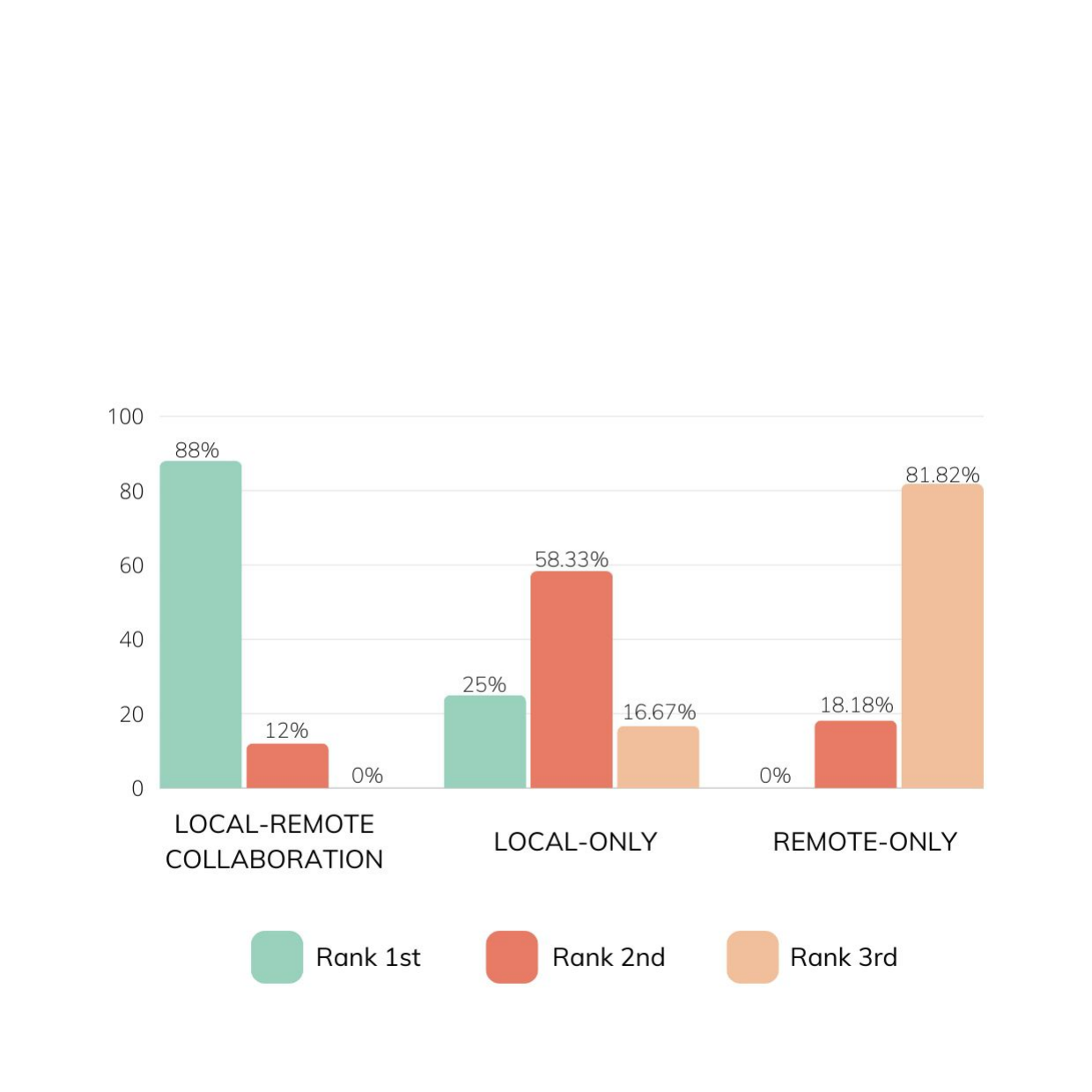}
    \caption{Vote Distribution of different LLM deployment strategies in Personal LLM strategies \citep{Personalllm}}
    \label{Personalllm}
\end{figure}

Edge-cloud (local-remote) collaborated deployment of LLM is preferred, while existing cloud-only (remote-only) (e.g., ChatGPT) is not a widely acceptable solution. As shown in Figure \ref{Personalllm}, 88$\%$ of participants prefer an edge-cloud collaborated architecture, 58.33$\%$ of them support local deployment, and 81.82$\%$ of them are not satisfied with the existing cloud-only solutions. Their main concerns are 1) the high latency of remote LLM service, 2) the risk of transmitting personal data to the cloud, and 3) the cost of cloud-based LLM services \citep{Personalllm}. 

Although cloud-based LLMs offer powerful capabilities, they come with certain drawbacks, including potential latency issues \citep{wang2024towards} and data concerns due to their dependency on networks. Hence, the concept of on-device deployment through edge computing has emerged to reduce latency and safeguard user data \citep{llamacpp}. Processing occurs locally, eliminating the need for data transmission. Moreover, the proliferation of customized hardware accelerators on mobile devices has made it feasible to run large LLMs with billions of parameters directly on devices.

On-device inference provides a compelling case for reducing latency because it allows models to run directly on the user's device without sending data to a cloud server. This approach is particularly beneficial for applications that require real-time responses. In the case of GPT-4, which gets responses based on the cloud, each token is generated at a speed of about 200 ms, while common end-side models can already generate tokens faster than this \citep{GPT4speed}. 

The ability to run models offline reduces reliance on network connectivity, making applications more accessible in areas with poor network coverage or other offline environments. For example, Google's Gemini Nano-based TalkBack, a feature that uses multimodal capabilities to recognize image content to provide voice broadcasts to people with disabilities, can work properly even when completely offline \citep{Talkback}. On-device reasoning also optimizes the use of limited computing resources through techniques such as model quantization, allowing language models to run efficiently even on devices with limited memory.

The deployment of LLMs on mobile devices is further facilitated by user-friendly interfaces that abstract away the complexities of AI, making the technology accessible to users without specialized knowledge. Moreover, these applications are not just limited to text generation but can extend their functionality to interact with device features, such as making calls, conducting web searches, and managing calendar events, through innovative text-to-actions features.

\subsection{The Performance Indicator of On-Device LLMs}

Latency is the time it takes from the user inputting a request to the system starting to respond. It usually refers to the time from when the model receives the input text to when it starts generating the first output. We generally use TTFT (Time-to-First-Token) to measure this metric \citep{hu2024inference, agrawal2024taming, agrawal2024metron}.

Inference speed refers to the speed at which the LLM makes an autoregression prediction of the next token based on all the previous tokens seen so far. However, in addition to the initial prompt decoding, inferring the next token also requires the logic of decoding one token at a time. This is because each new token depends on the previous token, and the previous token cannot be known in advance. This step takes up the most time in the reasoning of the large language model. Because of this, the speed of this step will mainly determine whether the user dialogue model is smooth, thus directly affecting the user experience \citep{ccoplu2023performance, cai2024medusa, zheng2024response}.

The size of RAM/VRAM used is also one of the performance indicators of language models operation. Due to the operation mechanism of language models, they consume corresponding RAM according to the size of model parameters during inference. For example, it is impractical to deploy a model with 70B parameters on a personal office laptop. This is crucial for many edge devices with limited RAM size. Engineers must use various model compression technologies to minimize the memory occupied by language model inference \citep{kwon2023efficient, zhao2024galore, zhao2024llm}.

In addition, the storage space occupied by models and the energy consumed during inference, for example, will become important indicators on edge devices. These indicators are particularly critical to whether LLMs can run on edge devices and how long they can run. In most cases, LLMs inference will put the processor into a fully loaded working state. If the operation time is too long, it will seriously consume the battery of the mobile device, thus bringing new problems. For example, a 7B parameter LLM inference will consume about 0.7J per token. For an iPhone with a battery capacity of about 50kJ, this means that the conversation with the model can only last for two hours at most. This does not take into account other issues such as device heating caused by model inference \citep{MobileLLM, stojkovic2024towards, jiang2024preventing}.

\section{Efficient Architectures for On-Device LLMs}
\label{sec: arch}


\subsection{Architectural Design Principles and Innovations for On-Device LLMs}

Designing language models for on-device deployment involves several architectural principles and innovations aimed at overcoming the resource constraints typical of mobile and edge devices. Key strategies include 1) parameter sharing \citep{ParameterSharing, ParameterSharing1}, which involves reusing weights across different parts of the model to reduce the overall parameter count; 2) modular architectures \citep{ning2023skeleton, ostapenko2024towards, jetmoe}, which break down the LLM into smaller, independent components or modules that can be processed separately or in parallel; and 3) compact representations, which focus on reducing the memory footprint of LLMs through techniques like quantization and weight pruning \citep{MobileLLM, EdgeShard, LLMcad}.
To provide a comprehensive comparison of these architectures, we consider their performance, computational efficiency, and memory requirements, which are summarized on Table \ref{table1}.







\begin{table*}[ht]
\centering
\caption{Comparative Analysis of State-of-the-Art On-Device LLM Architectures \label{table1}}
\footnotesize
\begin{tabular}{p{2.25cm}||p{3cm}|p{3.25cm}|p{3.7cm}}
\hline
\textbf{Model} & \textbf{Performance} & \textbf{Computational Efficiency} & \textbf{Memory Requirements} \\ \hline
\textbf{MobileLLM} \citep{MobileLLM}& High accuracy, optimized for sub-billion parameter models & Embedding sharing, grouped-query attention & Reduced model size due to deep and thin structures \\ \hline
\textbf{EdgeShard} \citep{EdgeShard} & Up to 50\% latency reduction, 2× throughput improvement & Collaborative edge-cloud computing, optimal shard placement & Distributed model components reduce individual device load \\ \hline
\textbf{LLMCad} \citep{LLMcad} & Up to 9.3× speedup in token generation & Generate-then-verify, token tree generation & Smaller LLM for token generation, larger LLM for verification \\ \hline
\textbf{Any-Precision LLM} \citep{Any-Precision} & Supports multiple precisions efficiently & Post-training quantization, memory-efficient design & Substantial memory savings with versatile model precisions \\ \hline
\textbf{Breakthrough Memory} \citep{sumsung} & Up to 4.5× performance improvement & PIM and PNM technologies enhance memory processing & Enhanced memory bandwidth and capacity \\ \hline
\textbf{MELTing Point} \citep{MELTing} & Provides systematic performance evaluation & Analyzes impacts of quantization, efficient model evaluation & Evaluates memory and computational efficiency trade-offs \\ \hline
\textbf{LLMaaS on MD \citep{LLMaaS}} & Reduces context switching latency significantly & Stateful execution, fine-grained KV cache compression & Efficient memory management with tolerance-aware compression and swapping \\ \hline
\textbf{LocMoE} \citep{locmoe} & Reduces training time per epoch by up to 22.24\% & Orthogonal gating weights, locality-based expert regularization & Minimizes communication overhead with group-wise All-to-All and recompute pipeline \\ \hline
\textbf{EdgeMoE} \citep{Edgemoe} & Significant performance improvements on edge devices & Expert-wise bitwidth adaptation, preloading experts & Efficient memory management through expert-by-expert computation reordering \\ \hline
\textbf{JetMoE \citep{jetmoe}} & Outperforms Llama2-7B and 13B-Chat with fewer parameters & Reduces inference computation by ~70\% using sparse activation & 8B total parameters, only 2B activated per input token \\ \hline
\end{tabular}
\end{table*}

\subsection{Model Compression and Parameter Sharing}

Efficient deployment of LLMs on resource-constrained devices such as smartphones and edge devices often requires reducing the model size without significantly sacrificing performance. Model compression and parameter-sharing techniques play a critical role in achieving this balance. This subsection reviews key research works that focus on optimizing sub-billion parameter LLMs through innovative compression and parameter-sharing methods.

\cite{AWQ} introduces a novel weight-only quantization method that focuses on the significance of weights in LLMs. AWQ protects a small fraction of crucial weights (0.1\%-1\%), reducing quantization loss and preserving the generalization ability of LLMs across different domains and modalities. Unlike traditional methods, AWQ does not require backpropagation or reconstruction, thus maintaining efficiency and performance. The proposed TinyChat inference framework implements AWQ, achieving significant speedup (up to 3×) over traditional FP16 implementations on both desktop and mobile GPUs.

MobileLLM addresses the need for efficient LLMs on mobile devices by proposing a deep and thin architecture optimized for sub-billion parameter counts \citep{MobileLLM}. This approach challenges the common belief that wider models are better, demonstrating that deep and thin structures can capture complex patterns effectively. Key techniques include embedding sharing, grouped-query attention, and block-wise immediate weight sharing. MobileLLM achieves significant accuracy improvements over previous state-of-the-art models (e.g., 2.7\% and 4.3\% accuracy boost over 125M and 350M models, respectively). The enhanced version, MobileLLM-LS, further increases accuracy while maintaining a small model size, making it ideal for on-device applications.

Both AWQ and MobileLLM showcase the potential of model compression 
 and parameter-sharing techniques in making LLMs feasible for deployment on mobile and edge devices. AWQ focuses on weight quantization to reduce model size and improve inference speed, while MobileLLM emphasizes architectural optimizations and weight sharing to create efficient sub-billion parameter models. These innovations are crucial for enhancing the performance and accessibility of LLMs in resource-constrained environments, enabling advanced AI capabilities on personal devices without compromising accuracy or efficiency.

\subsection{Collaborative and Hierarchical Model Approaches}
Deploying language models on resource-constrained devices faces significant challenges, such as limited memory and computational power. Collaborative and hierarchical model approaches offer innovative solutions to overcome these limitations by distributing the computational load and leveraging multiple models with varying capabilities. This subsection reviews key research works that implement collaborative and hierarchical strategies to enhance the efficiency and scalability of on-device LLMs.

EdgeShard introduces the EdgeShard framework, which partitions large LLMs into smaller segments (shards) and strategically distributes them across edge devices and cloud servers \citep{EdgeShard}. This method reduces latency and improves throughput by utilizing the computational power of multiple devices simultaneously. A dynamic programming algorithm optimizes shard placement, balancing the computational load and minimizing communication overhead. Experimental results show significant improvements in latency reduction (up to 50\%) and throughput enhancement (up to 2×) compared to traditional cloud-based methods.

LLMCad presents a novel inference engine that combines a smaller, memory-resident LLM with a larger, more accurate LLM \citep{LLMcad}. The smaller LLM generates candidate tokens, while the larger LLM verifies and corrects these tokens. This "generate-then-verify" approach leverages the efficiency of the smaller model and maintains the accuracy of the larger model. LLMCad introduces several techniques, including token tree generation and verification, self-adaptive fallback strategy, and speculative generation pipeline. These innovations enable LLMCad to achieve up to 9.3× speedup in token generation without compromising accuracy, making it suitable for real-time applications on mobile devices.

WDMoE proposed a new paradigm for deploying LLMs in a wireless communication system \citep{WDMoE}. By performing MoE Layer Decomposition, the gating network at the base station is deployed, and expert networks are distributed across mobile devices to optimize performance and reduce latency. In addition, the Expert Selection Policy is proposed to Dynamically adjust expert selection based on wireless channel conditions to ensure optimal performance.

Collaborative and hierarchical model approaches, such as those proposed in EdgeShard and LLMCad, offer effective solutions to the challenges of deploying LLMs on resource-constrained devices. By distributing the computational load across multiple devices and using smaller models for preliminary tasks, these methods enhance the scalability and efficiency of LLM inference. The EdgeShard framework demonstrates the benefits of collaborative edge-cloud computing, while LLMCad showcases the potential of hierarchical model collaboration in maintaining accuracy and improving inference speed. These approaches are crucial for enabling advanced AI capabilities on mobile and edge devices, providing real-time performance and efficient resource utilization.

\subsection{Memory and Computational Efficiency}

Efficient memory and computational resource utilization are critical for deploying large language models (LLMs) on mobile and edge devices. Various techniques and innovations aim to optimize the use of limited resources to ensure that LLMs can perform effectively without overwhelming the device's capabilities. This subsection reviews key research works focusing on enhancing memory and computational efficiency for on-device LLMs.

The researchers from Samsung Electronics proposes innovative memory solutions to address the memory bottlenecks in LLM deployment \citep{sumsung}. The authors introduce Processing-in-Memory (PIM) and Processing-near-Memory (PNM) technologies:

Aquabolt-XL \citep{Aquabolt-XL} and LPDDR-PIM \citep{LPDDR-PIM}: These PIM devices embed logic within the memory core, boosting internal memory bandwidth and supporting high-performance computing tasks, including LLM acceleration.
AXDIMM \citep{axdimm} and CXL-PNM: These PNM solutions place computational logic near the memory core, enhancing memory bandwidth and capacity. CXL-PNM integrates computational logic into the CXL memory controller, significantly improving memory capacity and performance.
Experimental results show that these memory solutions achieve up to 4.5× performance improvement and 71\% energy reduction compared to traditional memory architectures, making them highly suitable for LLM inference on resource-constrained devices.

MELTing Point introduces the MELT infrastructure, designed to facilitate the execution and benchmarking of LLMs on mobile devices \citep{MELTing}. The MELT framework supports Android, iOS, and Nvidia Jetson devices and provides detailed performance and energy metrics. MELT systematically evaluates on-device LLM execution, providing insights into performance, energy efficiency, and memory usage across various models.
The paper examines the impact of model quantization on performance and accuracy, demonstrating that while quantization reduces memory requirements, it incurs an accuracy cost.
The results highlight the importance of balancing memory and computational efficiency with performance to make LLMs viable for mobile applications.

Memory and computational efficiency are paramount for deploying LLMs on mobile and edge devices. The research works reviewed in this subsection present innovative solutions to overcome the memory wall and optimize resource usage. Samsung's memory solutions, such as PIM and PNM, significantly enhance memory bandwidth and capacity, enabling efficient LLM inference. The MELT infrastructure provides a comprehensive evaluation framework, offering valuable insights into the trade-offs between performance, energy efficiency, and memory usage. These advancements are crucial for ensuring that LLMs can operate effectively on resource-constrained devices, paving the way for more practical and efficient AI applications in mobile and edge environments.

\subsection{Mixture-of-Experts (MoE) Architectures}
Mixture-of-Experts (MoE) architectures offer a promising approach for deploying LLMs on edge devices by leveraging sparse activation and dynamic routing to improve efficiency. This subsection reviews key research works focusing on MoE-based models designed to optimize performance and resource utilization in on-device deployments.

EdgeMoE introduces a framework designed to efficiently execute MoE models on edge devices \citep{Edgemoe}. The authors proposed the Expert-wise Bitwidth Adaptation to reduce the size of expert weights with minimal accuracy loss using per-channel linear quantization. By utilizing novel expert management methods, they preload expert weights into the compute-I/O pipeline to reduce I/O swapping overhead.
Experimental Results demonstrate significant memory savings and performance improvements compared to baseline solutions, achieving up to 2.78× speedup in inference.

LocMoE introduces a routing strategy and communication optimization scheme to improve the efficiency of training MoE-based LLMs \citep{locmoe}. The Orthogonal Gating Weights method is employed to reduce computational costs and facilitate explicit routing decisions. Moreover, they introduced Locality-Based Expert Regularization to Encourage local experts to compete, reducing communication time and avoiding under-training.  They also included Group-Wise All-to-All and Communication Overlap to optimizes All-to-All operations by overlapping computation with communication to mask delays.

\cite{LLMaaS} proposed the LLMaaS paradigm, integrating large language models as a system service on mobile devices. In their proposed design, Stateful Execution allows the system to maintain persistent states (KV cache) across multiple invocations to improve performance.
The Unified Interface helps reduce memory usage by exposing LLMs and their infrastructure as a system feature to mobile apps. They also introducd techniques like chunk-wise KV cache compression and swapping to minimize context-switching overhead.

JetMoE presents an efficient approach to large language model training using a Sparsely-gated Mixture-of-Experts (SMoE) architecture \citep{jetmoe}. The authors apply sparse activation to both attention and feed-forward layers, significantly reducing computational costs while maintaining high performance. JetMoE-8B, trained with less than \$0.1 million using 1.25T tokens and 30,000 H100 GPU hours, outperforms Llama2-7B, while JetMoE-8B-Chat surpasses Llama2-13B-Chat. The model's 8B total parameters with only 2B activated per input token reduces inference computation by about 70\% compared to Llama2-7B. 

MoE architectures offer innovative solutions to the challenges of deploying LLMs on edge devices. These approaches leverage sparse activation and dynamic routing to improve computational efficiency and resource utilization. 

\subsection{General Efficiency and Performance Improvements}
Achieving efficient deployment of LLMs on edge devices involves a range of strategies aimed at improving overall performance while managing computational and memory constraints. This subsection reviews key research works that introduce innovative approaches to enhance the efficiency and effectiveness of on-device LLMs.

Any-Precision LLM proposes a novel method to deploy various LLMs with different precisions in a memory-efficient manner \citep{Any-Precision}. Any-Precision model extends any-precision deep neural networks to LLMs, allowing a single n-bit quantized model to support multiple lower bit-width models down to 3 bits. This reduces memory usage without significant performance loss.
Post-training quantization (PTQ) creates low-bit models and incrementally upscales them to higher bit widths. This avoids multiple training phases for each precision, saving time and resources.
A new software engine optimized for any-precision support manages memory bandwidth and improves serving efficiency, ensuring practical deployment of LLMs on edge devices.
The experimental results demonstrate substantial memory savings and improved serving efficiency, making any-precision LLMs suitable for a variety of on-device applications.

\cite{viability} explores the use of LLMs in software-hardware co-design to optimize the development of compute-in-memory (CiM) deep neural network (DNN) accelerators. The LCDA framework integrates LLMs into the design process of hardware and software, leveraging their extensive training on diverse datasets to speed up co-design.
 By incorporating heuristic knowledge from pre-trained LLMs, the framework bypasses the cold start problem, enabling faster convergence to optimal solutions.
The framework shows a 25x speedup in the design process compared to state-of-the-art methods while maintaining comparable performance levels in designing efficient DNN models and hardware architectures.
This approach highlights the potential of LLMs to enhance the co-design process, improving both software and hardware efficiency for advanced AI applications.

General efficiency and performance improvements are crucial for the practical deployment of LLMs on edge devices. The research works reviewed in this subsection introduce innovative methods to enhance memory efficiency, computational speed, and overall performance. The Any-Precision LLM approach offers a flexible and memory-efficient solution for deploying multiple LLMs with different precisions, while the LCDA framework demonstrates the benefits of integrating LLMs into the co-design process for optimizing both software and hardware. These advancements contribute to making LLMs more accessible and effective in resource-constrained environments, enabling a broader range of AI applications on mobile and edge devices.

\section{Model Compression and Optimization Techniques for On-Device LLMs}
\label{sec: comp}
In the realm of LLMs, optimizing computational efficiency while preserving performance is crucial, particularly for deployment on edge devices. This section examines four key model compression techniques: quantization, pruning, knowledge distillation, and low-rank factorization. These methods enhance the operational efficiency of LLMs, ensuring their viability for on-device applications by balancing performance, memory footprint, and inference speed.

\subsection{Quantization}
Quantization in neural networks refers to the process of transforming high-precision (floating-point) weights and activations into lower bit-widths (integers). This technique substantially reduces the model size and computational demands, enabling faster inference and decreased memory consumption while preserving accuracy.

\begin{enumerate}
    \itemsep0em
    \parsep0em
    \item \textbf{Post-training quantization (PTQ)} : PTQ is applied after model training, requiring no retraining and thus being faster and less resource-intensive than quantization-aware training (QAT). There are a few notable PTQ methods. GPTQ \citep{frantar2022gptq} utilizes second-order information for error compensation, effectively reducing bit width to 3 or 4 bits per weight. This method maintains high accuracy with minimal perplexity increase, enabling language models like OPT-175B to run on a single high-end GPU. Activation-aware Weight Quantization (AWQ) \citep{lin2024awq} is based on the observation that a small fraction (0.1\%-1\%) of weights are crucial for LLMs' performance. By selectively skipping quantization of these salient weights, AWQ significantly reduces quantization loss. 
    
    \begin{enumerate}
        \item \textbf{Weight-only quantization} : In weight-only quantization, only the weights of the neural network are quantized. This approach simplifies the quantization process and can be particularly effective when activations do not vary significantly in range or when computational resources are severely limited.
        \item \textbf{Weight-activation co-quantization} : Both weights and activations are quantized, enhancing reduction in computational complexity. This method is advantageous in hardware implementations due to efficient matrix multiplication \citep{dettmers2022gpt3}, vital in neural computations. BitNet b1.58 \citep{ma2024era} uses ternary quantization {-1, 0, 1} for each parameter, significantly enhancing latency, memory, throughput, and energy consumption metrics.
    \end{enumerate}
    \item \textbf{Quantization-aware training (QAT)} : QAT incorporates quantization directly into the training process, allowing the model to accommodate the reduced precision constraints inherently. This integration generally yields higher accuracy post-quantization, as the model proactively learns to compensate for potential quantization errors during its training phase.
\end{enumerate}

\subsection{Pruning}
Pruning in neural networks involves selectively removing weights or neurons to reduce complexity and enhance computational efficiency without significantly compromising performance. This process targets the less crucial components of a model, focusing on efficiency and functional integrity.

\begin{enumerate}
    \item \textbf{Structured Pruning}: This approach removes entire subsets of parameters like layers, channels, or filters, which is beneficial for hardware optimization due to more regular memory access patterns and simplified computations. The `LLM-Pruner' \citep{kaddour2023challenges} employs structured pruning to eliminate non-essential groups based on gradient data, thus maintaining critical functionalities. It also facilitates performance recovery through techniques such as LoRA, allowing efficient restoration with minimal data.

    \item \textbf{Unstructured Pruning}: Unlike structured pruning, unstructured pruning removes individual weights across the model, offering finer granularity and potentially higher compression rates \citep{li2023model}. However, this method typically results in sparse matrices, which can be less compatible with traditional hardware architectures, compromising computational efficiency. It is most suitable where maximum compression is needed without constraints on structural preservation.

    \item \textbf{Contextual Pruning}: This advanced method prunes based on the operational context of the model, targeting weights or neurons that are only relevant under specific conditions or for particular tasks. Contextual pruning ensures that reductions align dynamically with the model’s operational needs, thereby preserving performance where it matters most.
\end{enumerate}

\subsection{Knowledge Distillation}

Knowledge Distillation (KD) is a technique for transferring knowledge from a large, computationally intensive model (teacher) to a smaller, more efficient model (student). This method is crucial for condensing the capabilities of large language models (LLMs) into more manageable forms without significantly impacting performance.

\begin{enumerate}
    \item \textbf{Black-box Knowledge Distillation}: This approach involves the student model learning solely from the outputs of the teacher model, without access to its internal mechanics or parameters. It is particularly advantageous when the teacher model's details are proprietary or when the architectures of the teacher and student models differ markedly. For instance, \cite{gu2023minillm} demonstrated that black-box KD could effectively train models using only the output data from LLM APIs like ChatGPT. The student model trains to emulate the teacher's output distribution based on input-output pairs, a process that, while effective, limits learning to external behaviors without tapping into the teacher's deeper internal states.

    \item \textbf{White-box Knowledge Distillation}: In contrast, White-box Knowledge Distillation allows the student model to access the internal states and workings of the teacher, facilitating a deeper and more precise learning process. This method enables the student to mimic not just the outputs but also the internal state distributions of the teacher, enhancing learning efficacy and depth. The increased access to the teacher's detailed workings helps guide the student's learning, resulting in more accurate and robust models. However, this technique requires a careful alignment of model architectures to ensure effective knowledge transfer and is generally more complex to implement.
\end{enumerate}

\subsection{Low-Rank Factorization}
Low-Rank Factorization (LRF) is a technique utilized to decompose matrices into smaller components, significantly reducing computational complexity without substantially impacting model accuracy. Leveraging the inherent low-rank structure prevalent in many real-world matrices, LRF facilitates the approximation of these matrices by products of low-rank factors, which has proven indispensable in applications such as image processing, dimensionality reduction in machine learning models, and data compression \citep{saha2023matrix}. This methodology not only maintains essential data characteristics but also ensures efficient storage and processing, highlighting its crucial role in modern computational tasks. Further extending its application, a study by \cite{yao2024exploring} integrates LRF with Post-training Quantization (PTQ) in Large Language Models. This innovative approach, termed Low-Rank Compensation (LoRC), enhances model efficiency by significantly reducing model size and preserving accuracy, effectively mitigating the detrimental effects of activation quantization. This synthesis of LRF and PTQ demonstrates a significant advancement in optimizing computational efficiency while maintaining performance integrity in complex models.

\section{Hardware Acceleration and Deployment Strategies}
\label{sec: hardware}

Hardware accelerators such as GPUs, TPUs, and specialized AI chips play a crucial role in enabling efficient on-device inference of LLMs by offering substantial computational capabilities and high memory bandwidth. The selection between GPUs, TPUs, FPGAs, and other AI-specific chips involves careful consideration of trade-offs involving performance, power consumption, and cost. For instance, GPUs are favored for their parallel processing prowess, TPUs for their specialized matrix operations, and FPGAs for their customizable hardware tailored to specific tasks, which can be more power-efficient. Software-hardware co-design approaches, including quantization-aware training and model compression, further enhance efficiency, making LLMs feasible on a range of devices from high-power servers to low-power edge devices. Optimization strategies like parameter sharing and advanced memory management techniques are vital for reducing the footprint of LLMs, ensuring faster and more cost-effective deployments across diverse computing environments. These strategies collectively improve the deployment and execution of LLMs, catering to various application needs and hardware constraints.

\subsection{Popular On-Device LLMs Framework}
Deployment strategies for LLMs can vary significantly depending on the use case and the available infrastructure, ranging from fully cloud-based solutions to edge-only deployments.

\begin{enumerate}
    \itemsep0em
    \parsep0em
    \item \textbf{Edge-only}
    \begin{enumerate}
        \item \textbf{Llama.cpp}
        \begin{itemize}
            \item \textbf{Description}: Llama.cpp \citep{llamacpp} is a C/C++ library designed for efficient inference of large language models on a broad range of hardware platforms. It supports integer quantization, GPU acceleration, and CPU+GPU hybrid inference.
            \item \textbf{Training}: Supports fine-tuning LORA adapters on-device.
            \item \textbf{Inference}: Supports CPU and CPU+GPU hybrid inference across ARM and x86 architectures.
        \end{itemize}
        \item \textbf{MNN}
        \begin{itemize}
            \item \textbf{Description}: MNN \citep{mnn} leverages Mobile Neural Network technology for efficient LLM inference on various platforms, optimized for mobile devices with dynamic inputs and multimodal interactions.
            \item \textbf{Training}: Supports full-sized fine-tuning and LORA fine-tuning on-device.
            \item \textbf{Inference}: Supports model deployment for ONNX and MNN formats across diverse backends including CPU, CUDA, and OpenCL.
        \end{itemize}
        \item \textbf{PowerInfer}
        \begin{itemize}
            \item \textbf{Description}: PowerInfer  \citep{song2023powerinfer} and PowerInfer2 \citep{xue2024powerinfer} is a high-speed inference engine optimized for deploying LLMs on PCs with consumer-grade GPUs, utilizing a locality-centric design.
            \item \textbf{Training}: No built-in training capabilities.
            \item \textbf{Inference}: Supports various computing platforms including x86-64 CPUs and Apple M Chips, optimized for Windows and Linux.
        \end{itemize}
        \item \textbf{ExecuTorch}
        \begin{itemize}
            \item \textbf{Description}: ExecuTorch \citep{executorch2024} is part of the PyTorch Edge ecosystem, designed for deploying PyTorch models efficiently on edge devices like mobile phones and wearables.
            \item \textbf{Training}: No built-in training capabilities.
            \item \textbf{Inference}: Leverages full hardware capabilities like CPUs, NPUs, and DSPs across various computing platforms.
        \end{itemize}
        \item \textbf{MediaPipe}
        \begin{itemize}
            \item \textbf{Description}: Developed by Google, MediaPipe \citep{mediapipe2024} is a framework for building and deploying multimodal machine learning pipelines involving video, audio, and other time-series data.
            \item \textbf{Training}: No built-in training capabilities.
            \item \textbf{Inference}: Supports multiple platforms including Android, iOS, macOS, Windows, and Linux, leveraging CPU and GPU resources.
        \end{itemize}
    \end{enumerate}
    \item \textbf{Edge-cloud}
    \begin{enumerate}
        \item \textbf{MLC-LLM}
        \begin{itemize}
            \item \textbf{Description}: MLC-LLM \citep{mlc-llm} is a machine learning compiler and high-performance deployment engine, supporting universal LLM deployment on edge devices and in cloud environments.
            \item \textbf{Training}: No built-in training capabilities.
            \item \textbf{Inference}: Supports inference on various platforms including CPUs and GPUs across ARM and x86 architectures.
        \end{itemize}
        \item \textbf{VLLM}
        \begin{itemize}
            \item \textbf{Description}: VLLM \citep{vllm2024} is optimized for edge-cloud environments, supporting advanced quantization methods for efficient key and value memory management during inference.
            \item \textbf{Training}: No built-in training capabilities.
            \item \textbf{Inference}: Supports multiple GPU platforms and integrates with Vulkan, CUDA, Metal, and WebGPU technologies.
        \end{itemize}
        \item \textbf{OpenLLM by BentoML}
        \begin{itemize}
            \item \textbf{Description}: OpenLLM \citep{openllm} enables the deployment of various open-source LLMs as OpenAI-compatible API endpoints, optimized for high throughput and streamlined cloud deployment.
            \item \textbf{Training}: No built-in training capabilities.
            \item \textbf{Inference}: Compatible with various model architectures and backend implementations for efficient deployment in production settings.
        \end{itemize}
    \end{enumerate}
\end{enumerate}

\subsection{Hardware Acceleration}
The continuous advancement in hardware technologies significantly impacts the deployment and performance of on-device LLMs.
\begin{enumerate}
    \itemsep0em
    \parsep0em
    
    \item \textbf{GPU}: Graphics Processing Units (GPUs) have become the standard for training and accelerating large language models due to their massive parallelism and high memory bandwidth. NVIDIA's Tensor Cores, introduced in the Volta architecture and improved in subsequent generations, offer specialized hardware for mixed-precision matrix multiply-accumulate operations, which are crucial for transformer-based models. Recent advancements like NVIDIA's A100 GPU with 80GB HBM2e memory enable training of models with billions of parameters on a single device. Techniques such as tensor parallelism and pipeline parallelism, implemented in frameworks like Megatron-LM, allow efficient scaling of LLMs across multiple GPUs. The use of mixed-precision training, particularly FP16 and BF16 formats, significantly reduces memory footprint and increases computational throughput on modern GPUs.
    
    \item \textbf{NPU}: Neural Processing Units (NPUs), also known as AI accelerators, are specialized chips designed for machine learning workloads. Google's Tensor Processing Units (TPUs) are a prominent example, with the latest v4 offering 275 TFLOPS of BF16 performance per chip. TPUs utilize a systolic array architecture for efficient matrix multiplications, which is particularly well-suited for transformer layers in LLMs. The TPU Pod configuration allows scaling to thousands of chips, enabling training of models like GPT-3 and PaLM. Huawei's Ascend AI processors and Apple's Neural Engine are other examples of NPUs that offer on-device acceleration for inference of smaller LLMs, utilizing techniques like quantization and pruning to reduce model size and computational requirements.
    
    \item \textbf{FPGA}: Field-Programmable Gate Arrays (FPGAs) offer a flexible hardware platform for accelerating LLMs, particularly for inference. Recent work has demonstrated efficient implementations of transformer layers on FPGAs, utilizing techniques such as sparse matrix multiplication and quantization. For example, Microsoft's Project Brainwave uses Intel Stratix 10 FPGAs to accelerate BERT inference, achieving low latency and high throughput. FPGAs excel in energy efficiency and can be optimized for specific model architectures, making them suitable for edge deployment of smaller LLMs. However, their lower computational density compared to GPUs and ASICs limits their application in training large-scale models.
\end{enumerate}

\section{Examples and Applications}
\label{sec: eg}
In the past years, the rapid development of artificial intelligence technology and the continuous upgrade of mobile device hardware have made the deployment of large language models on edge devices a reality. Smartphones are one of the most commonly used devices in people's daily lives, and the language models on them are particularly eye-catching. At present, the world's major mobile phone brand manufacturers have developed and released a number of advanced models that are deployed on the device side or adopt device-cloud collaboration strategies, as displayed in Table \ref{manufacturers}. These models not only mark a major leap forward in mobile computing but also bring users a series of advantages that traditional cloud deployments cannot match.

\begin{table*}[ht]
\centering
\caption{State-of-the-Art On-Device LLM released by mobile phone manufacturers}
\begin{tabular}{c|c|c|c|c}
\hline
\textbf{Year} & \textbf{MODEL NAME} & \textbf{Model Size} & \textbf{Edge} & \textbf{Cloud} \\
\hline
2023 & Google Gemini Nano & 7B & $\surd$ & \quad\\
2023 & OPPO AndesGPT & 7B & $\surd$ & $\surd$\\
2024 & Honor MagicLM & 7B & $\surd$ &  \quad\\
2024 & VIVO BlueLM & 7B & $\surd$  & $\surd$\\
2024 & XiaoMi MiLM & 6B & $\surd$ & \quad \\
2024 & Apple OpenELM & 1.1B & $\surd$ & $\surd$ \\
\hline
\end{tabular}
\label{manufacturers}
\end{table*}

\subsection{Examples of on-device language models}
\begin{enumerate}
    \item\textbf{Gemini Nano:}
Mobile operating system will expose and LLM and its inference infrastructure as a system feature to mobile apps, like the location or notification services. User can access AI core through Google AI Edge SDK. Inside of AI core, google provide a Gemini Nano model, which is smaller than other other Gemini models running inference in the cloud, but with faster speed and low inference. AI core is responsible for the distribution of Gemini Nano model so the memory can be managed well. Besides, AI core can perform at the best speed since it leverages on-device hardware to accelerate inference.
Gemini Nano model is trained by distilling from larger Gemini models. It
is 4-bit quantized for deployment and provides best-in-class performance \citep{team2023gemini}.


\item\textbf{Nexa AI Octopus series model}: A 2 billion parameter model running on edge device surpasses GPT-4 in accuracy and latency and reduces context length by 95$\%$. By tokenizing the names of core functions and fine-tuning the model using functional tokens, the model can understand the functionality of the software application and learn to map function descriptions to specific tokens. Deployment of the Octopus model on mobile devices demonstrated fast response times, completing function calls in 1.1 to 1.7 seconds for a typical query of 20 to 30 tokens, even on a standard Android phone \citep{chen2024octopus,chen2024octopus2,chen2024octopus3,chen2024octopus4}.

\item \textbf{Apple OpenELM and Ferret-v2}: Apple has developed OpenELM \citep{mehtaOpenELMEfficientLanguage2024}, a substantial large language model integrated within iOS to augment application functionalities, analogous to essential system services such as location tracking. OpenELM employs a layer-wise scaling architecture, efficiently deploying its 1.1 billion parameters to achieve a 2.36\% increase in accuracy compared to prior models, while requiring only half the pre-training tokens. Moreover, it is compatible with the MLX library, facilitating direct fine-tuning on Apple devices. In parallel, Ferret-v2 \citep{zhang2024ferretv2} marks a significant upgrade over its predecessor, incorporating features such as any-resolution grounding, multi-granularity visual encoding through the integration of a DINOv2 encoder, and a sophisticated three-stage training regimen. These enhancements markedly improve performance by advancing high-resolution image processing and enriching visual comprehension, thereby ensuring robust, on-device functionality for iOS users.

\item\textbf{Microsoft Phi series}: Microsoft's latest Phi-3-mini \citep{abdin2024phi3} a compact yet powerful 3.8 billion parameter language model, trained on an extensive 3.3 trillion token dataset. Despite its small size suitable for mobile deployment, Phi-3-mini delivers performance competitive with larger models like Mixtral 8x7B and GPT-3.5, achieving 69\% on MMLU and 8.38 on MT-bench. This model benefits from a unique training dataset, an expanded version of the one used for Phi-2, which combines heavily filtered publicly available web data with synthetic data, enhancing robustness, safety, and chat functionality. Additionally, we present initial results from our scaled models, Phi-3-small and Phi-3-medium, trained on 4.8 trillion tokens, with 7 billion and 14 billion parameters respectively, showing superior capabilities (75\% and 78\% on MMLU, and scores of 8.7 and 8.9 on MT-bench). Expanding further, we introduce Phi-3-vision, a 4.2 billion parameter model derived from Phi-3-mini, designed with enhanced reasoning abilities for both image and text prompts.

\item\textbf{MiniCPM}: 
The MiniCPM-Llama3-V 2.5, a recent addition to the open-source MiniCPM-V lineup crafted by the collaborative efforts of Tsinghua University and ModelBest, boasts a substantial parameter count of 8.5 billion \citep{MiniCPM}. This model has demonstrated exceptional performance across the OpenCompass assessment platform, which encompasses a wide array of 11 multimodal benchmarks. With a noteworthy average score of 65.1, MiniCPM-Llama3-V 2.5 has surpassed leading industry models, including GPT-4V-1106 at 63.5, Gemini Pro at 62.9, Claude 3, and Qwen-VL-Max, even though it possesses only a fraction of the parameters these models have.

In specific evaluations focusing on Optical Character Recognition (OCR) and scene text comprehension, MiniCPM-Llama3-V 2.5 has excelled, securing a score surpassing the 700-point mark on OCRBench, thereby outdoing its counterparts such as GPT-4 and Gemini Pro. Moreover, it has attained remarkable accuracy rates of 76.6\% on the TextVQA benchmark and an impressive 84.8\% on DocVQA, effectively establishing a new standard for the performance of open-source models in these domains.


\item\textbf{Gemma2-9B:} Gemma is a lightweight, state-of-the-art family of open models from Google. Gemma2 is Google's upgraded version of Gemma, available in two different sizes, 9B and 27B. For the 9B version, Gemma2 has a training data volume of 8 TB Tokens of web data, code and math data. The authors have taken a novel approach to combining attention, with one layer of sliding window attention and one layer of global attention. Techniques such as knowledge distillation, model merging, etc., were also used. Gemma2-9B model also performs well in its equivalent volume category, outperforming Llama 3-8B and other similar open models in several domains such as reasoning, math, and code. This model also has good compatibility with major AI frameworks such as HuggingFace, as well as Keras 3.0, vLLM, Gemma.cpp, and Llama.cpp \citep{Gemma2}.

\item\textbf{Qwen2-0.5B:} Qwen team, Alibaba Cloud has upgraded the Qwen model series to Qwen2 and brought the series to five sizes. Among them, Qwen2-0.5B is the one with the smallest number of parameters and a context length of 32K. In multiple tests, Qwen2-0.5B performs similarly to Gemma-2B and Phi-2 \citep{Qwen2}, but has a smaller number of parameters, which makes it possible to play a big role in the future of the smart home industry.
In addition, for the problem of short context length, the Qwen-Agent framework adopts the idea of Agentic RAG, which can extend the processing context to 1M, thus realizing long text understanding \citep{bai2023qwen}.
\end{enumerate}

\subsection{Applications of On-Device LLMs}
\begin{figure}[!htb]
    \centering
    \includegraphics[width=0.7\textwidth]{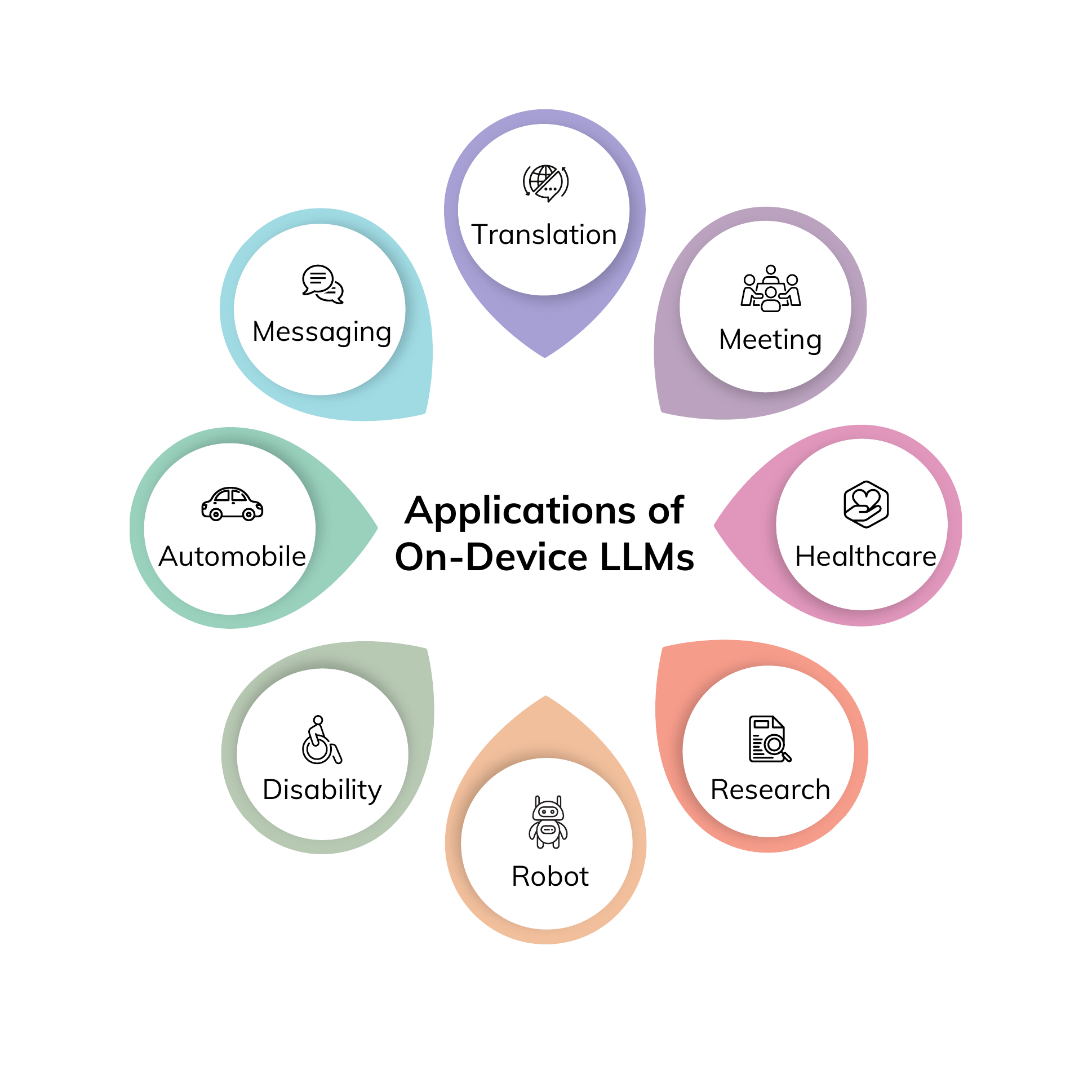}
    \caption{Different application domains of on-device LLMs}
    \label{app}
\end{figure}

On-device language models are ushering in a new era of intelligent, responsive, and personalized applications. By bringing the power of advanced natural language processing directly to end-user devices, these models are transforming how we interact with technology in our daily lives and professional endeavors. From instantaneous message suggestions to real-time language translation, from confidential medical consultations to cutting-edge autonomous vehicles, on-device LLMs are proving to be versatile tools with far-reaching implications. The following examples, as summarized in Figure \ref{app}, illustrate the breadth and depth of on-device LLM applications, showcasing how this technology is not only enhancing existing services but also enabling entirely new categories of intelligent, responsive, and secure applications across diverse domains.

\begin{enumerate}
    \item\textbf{Text Generating For Messaging:}
In the past, the quick reply function based on cloud LLM was limited by the generation speed and network latency, so it would be slow to generate reply for users. This is inefficient in fast-paced instant conversations. Thanks to on-device LLMs, Gboard (Keyboard app by Google) can use the Gemini Nano, an on-device LLM by Google \citep{Google2024Gboard}.  When it detects that the user is chatting online, Gemini Nano can quickly generate conversation-aware quick replies for the user to choose from based on the chat content. Because the language models used does not need to be connected to the Internet to wait for the server to respond, this function can reflect the true response speed.

\item\textbf{Translation:}
LLMs have been widely used in the field of language translation. This method can use terminology and style suitable for a specific field for translation, which is not possible with traditional machine translation methods. However, cloud-based LLMs still face problems such as slow response speed and the need to upload information.
On-device LLMs better solve these problems, with smaller parameters, faster response speed, and can also run in offline environments. This also provides data security for many scenarios.
In terms of translation quality, using small-size models does not significantly reduce the accuracy of translation. The token generation accuracy using the T5-small model is only 4\% lower than the T5-language models \citep{LLMcad}. In addition, faster response speed means that the on-device model will be more suitable for more immediate translation situations such as simultaneous interpretation.

\item\textbf{Meeting Summarizing:}
Distill-CLI, a cloud-based solution released by Amazon CTO, uses Anthropic's Claude 3 Sonnet model and Amazon Transcribe technology to generate real-time meeting summaries \citep{Distill}. Similar applications such as Plaud Note with GPT-4o model \citep{Plaud}, Zoom-IQ \citep{Zoom-IQ}, etc. However, the disadvantage of using cloud-based models is that subscription service fees will be incurred, as well as network latency problems caused by networking. By employing an on-device model, the data remains localized and does not require uploading to a cloud-based server.

\item\textbf{Healthcare application:}
Current medical models, like Med-Palm Multimodal \citep{tu2024towards} can combine and analyze patient statements, electronic record information, X-rays and other medical images to generate long-form responses with high accuracy. 
Edge deployment can help patients answer questions offline, thereby ensuring the emergency availability of the model and keeping the patient's condition localized. What is exciting is that models fine-tuned based on pre-trained models in professional medical fields have emerged, such as BioMistral-7B \citep{labrak2024biomistral}, HuatuoGPT-7B-II \citep{chen2023huatuogpt}, etc. These low-parameter models have the potential to be deployed on terminal devices.

\item\textbf{Scientific Research Support:}
Traditional research support LLMs like GatorTronGPT \citep{peng2023study} use large amount of certain professional data to train. This enables them to generate high-quality professional text, thereby accelerating the progress of science research, especially in research areas where data is scarce or sensitive.

After changing to on-device LLMs, it can reduce the hardware cost of using language models to assist scientific research tasks, obtain faster responses, and protect the confidentiality of scientific research information.


\item\textbf{Companion Robot:}
There are already some research cases that use language models to enhance the capabilities of robots or Internet of Thing (IoT) devices  \citep{ahn2022can, xu2024general}.  LLM's powerful planning and reasoning capabilities can decompose human instructions into a series of text subtasks, allowing robots to better understand natural language instructions \citep{zeng2023large}. For example, the Figure 01 robot based on Open AI's multimodal language models can communicate deeply with people and make independent decisions and actions based on the content of the conversation \citep{Figure01}.
With the rise of small-size models, robots that deploy on-device language models can outperform traditional cloud-based model robots in terms of corresponding generation speed. At the same time, the client-side model can ensure that the robot can still maintain its intelligent capabilities when offline.



\item\textbf{Disability Support:}
For visually impaired users, converting images into text is a very basic and important function. Currently, there are many on-device large multimodal models, like Octopus v3 \citep{chen2024octopus3}, MiniCPM-Llama3-V 2.5 \citep{MiniCPM} that can achieve this function by multimodel ability. With them, blind people can also easily know the picture and video information in the conversation.

Google is about to launch Talkback feature based on Gemini Nano, helping people who are blind or have low vision to describe what is happening in the image more richly and clearly \citep{Talkback}. Because Gemini Nano is a model deployed on the edge, these descriptions will appear quickly and work even without a network connection.

Similar capabilities can also be used for sign language recognition, and there are projects that use the ChatGPT model for sign language translation \citep{sincan2024using}. In comparison, the on-device model can generate text translations corresponding to sign language with lower latency and ensure its offline availability.


\item\textbf{Autonomous Vehicles:}
Using language models to drive autonomous cars may be an ideal future, but we already have examples of this today.
DriveVLM Dual is a system that combines autonomous driving technology with a large-scale visual language model (VLM) to improve the understanding of complex and long-tail scenes in urban environments. The system uses language to describe the driving environment and identify key objects in the scene. It gradually develops a plan from meta-action and decision descriptions to waypoints. DriveVLM surpasses existing state-of-the-art methods on both public benchmarks and the researchers' own benchmarks, especially in handling complex and dynamic scenes. Excitingly, DriveVLM can be deployed locally on the car, which also provides convenience for its immediate response \citep{tian2024drivevlm}.
\end{enumerate}

\section{Future Directions and Open Challenges}
\label{sec: directions}
\begin{figure}[bht]
    \centering
    \includegraphics[width=0.75\textwidth]{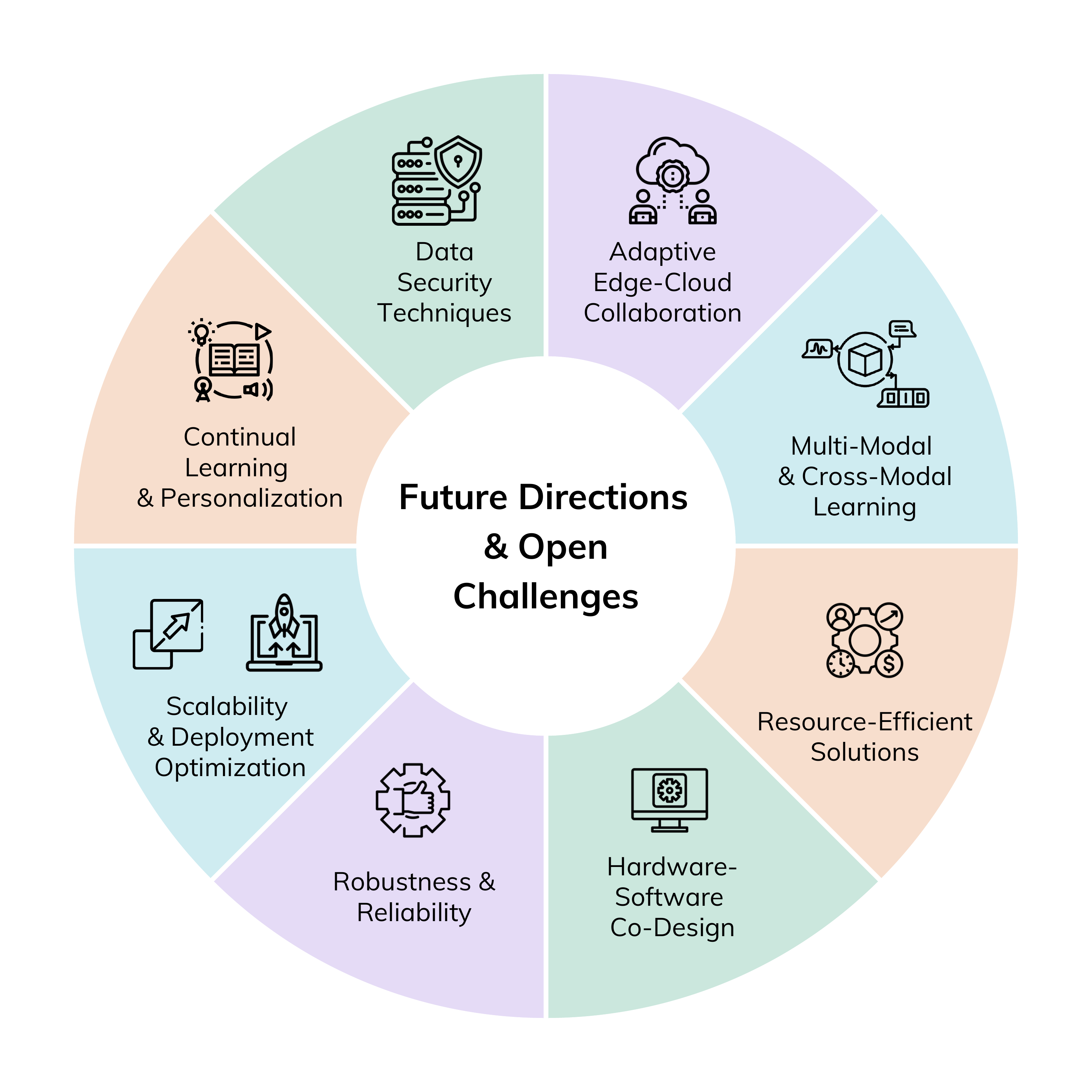}
    \caption{Future Directions and Open Challenges for on-device LLMs}
    \label{directions}
\end{figure}

As on-device LLMs continue to evolve, several vital areas emerge as promising future research and development directions. The field of on-device LLMs is rapidly advancing, driven by the increasing demand for 1) data security, 2) low-latency, and 3) personalized AI experiences on edge devices. This progress is exemplified by recent developments such as TinyLlama \citep{Tinyllama}, MobileVLM \citep{mobilevlm,mobilevlmv2}, and novel approaches like the OpenELM \citep{mehtaOpenELMEfficientLanguage2024}.
However, deploying LLMs on resource-constrained devices presents unique challenges that differ significantly from traditional cloud-based implementations. These challenges span multiple areas, including model compression, efficient inference, security, energy efficiency, and seamless integration with diverse hardware platforms. Moreover, the dynamic nature of edge environments and the need for continuous adaptation introduce additional complexities that must be considered.
We outline the most pressing challenges and opportunities in advancing the field of LLMs on-device. By identifying these key areas and stimulating innovation in developing more capable, efficient, and reliable on-device language models, we aim to provide insights for future research efforts. We should notice that the challenges and opportunities discussed here are interconnected: the progress in one area often has implications for others. Therefore, a holistic approach that considers the interplay between different aspects of on-device LLM deployment is essential for achieving significant advancements in the field.
We delve into the current state of research, identifying key challenges and proposing potential directions for future work, summarized in Fig. \ref{directions}. By addressing these challenges,  researchers and practitioners can push the boundaries of what is possible with on-device LLMs, ultimately leading to more intelligent, efficient, and user-centric computing experiences across various applications and domains.

\subsection{Data Security Techniques}
On-device language models may offer inherent data security advantages, since all the data can remain localized. Future work should focus on:
\begin{itemize}
    \item Developing efficient privacy techniques techniques, including query obfuscation \citep{yuan2024wip}, prompt tuning \citep{li2023privacy}, and advanced randomization techniques \citep{zhang2024no} that balance data security guarantees with model utility and computational constraints.
    \item Enhancing risk assessment and monitoring, by creating sophisticated benchmarking systems \citep{yuan2024wip}, implementing real-time monitoring \citep{das2024security}, and designing systems to detect and mitigate potential PII leakage during inference \citep{kim2024propile}.
    \item Optimizing model architectures and communication strategies, focusing on efficient model sharding \citep{yang2024pfid}, security-enhancing architectures \citep{yao2024survey}, and minimizing data transmission \citep{wang2023privatelora}.
    \item Addressing security challenges in collaborative and distributed learning scenarios, through secure multi-party computation \citep{das2024security}, data protection for long conversations \citep{yuan2024wip}, and extending frameworks like PFID to support a wider range of LLM architectures and tasks \citep{yang2024pfid}.
\end{itemize}

\subsection{Adaptive Edge-Cloud Collaboration}
As on-device language models continue to evolve, the synergy between edge computing and cloud infrastructure presents both opportunities and challenges. Future research in adaptive edge-cloud collaboration for on-device LLMs should explore:

\begin{itemize}
    \item Inventing advanced caching and request analysis techniques, including sophisticated vector database caching strategies, feature extraction models for diverse LLM requests \citep{velo}, and uncertainty-guided token sampling methods to optimize data transmission between edge devices and cloud servers \citep{wang2024cloud}.
    \item Designing intelligent scheduling and resource allocation algorithms, incorporating personalized inference scheduling \citep{velo}, adaptive resource allocation for heterogeneous infrastructures \citep{perllm}, and batch size-aware optimization techniques to efficiently distribute LLM components and workloads across edge-cloud environments \citep{EdgeShard}.
    \item Creating efficient knowledge transfer and model compression methods, such as adapter-based knowledge distillation for multimodal LLMs \citep{zhang2024edge}, dynamic quantization techniques for various LLM architectures, and adaptive weight update compression strategies to enable effective deployment of language models on resource-constrained devices \citep{wang2024cloud}.
    \item Improving performance optimization in collaborative systems by developing adaptive control mechanisms for token-level collaboration \citep{perllm}, efficient constraint satisfaction algorithms for real-time decision-making, and innovative techniques to reduce latency and improve pipeline execution in hybrid edge-cloud systems \citep{hao2024hybrid, EdgeShard}.
\end{itemize}

\subsection{Multi-Modal and Cross-Modal Learning}
As LLMs expand to incorporate multiple modalities, there is a growing need for efficient multi-modal architectures suitable for on-device deployment \citep{carreira2023revolutionizing, MobileLLM}. Key research directions include:

\begin{itemize}
    \item Developing efficient multi-modal processing and compression techniques, including advanced uncertainty-guided token sampling methods, dynamic weight update compression strategies for cloud-to-device model updates \citep{wang2024cloud, mm1}, and innovative approaches to efficiently combine multiple modalities like audio, text, and video for on-device models \citep{wagner2024multimodal}.
    \item Enhancing knowledge transfer and adaptation capabilities, such as exploring advanced adapter-based knowledge distillation methods for transferring knowledge from larger cloud models to smaller on-device models, improving few-shot and zero-shot capabilities across modalities \citep{chen2024gui, onellm, mm1}, and investigating hybrid approaches that combine generative and retrieval-based methods for multimodal content generation \citep{nextgpt}.
    \item Expanding modality support and improving multi-modal understanding, through the development of large-scale datasets for non-image modalities, design of new encoders for fine-grained multi-modal understanding of high-resolution images, long video sequences, and complex audio inputs \citep{onellm}, and incorporation of support for additional modalities and tasks like web pages, 3D vision, heat maps, and tables/figures \citep{nextgpt}.
    \item Advancing temporal and contextual processing abilities, by investigating longer context windows that incorporate features from previous interactions, developing sophisticated techniques for processing and understanding temporal and sequential information across modalities, and exploring tasks useful during interactions with virtual assistants, such as audio captioning and acoustic scene classification \citep{wagner2024multimodal}.
\end{itemize}

\subsection{Resource-Efficient Solutions}
The deployment of LLMs on edge devices raises concerns about energy consumption and environmental impact. Future research should prioritize:
\begin{itemize}
    \item Creating efficient model compression and execution algorithm: Develop advanced pruning, quantization, and knowledge distillation techniques for LLMs. Explore methods to optimize execution for larger-than-memory models. Investigate dynamic and adaptive inference techniques to adjust model complexity based on input and available resources \citep{beyond-efficiency}.
    \item Exploiting model sparsity: Investigating techniques to take advantage of the runtime activation sparsity of language models, where only a small portion of the model is activated for a given task. This could lead to significant reductions in inference time and memory footprint, enabling more efficient scaling of model sizes \citep{resource-efficient}.
    \item Developing energy-aware training and deployment strategies, including energy-efficient algorithms and runtime optimizations \citep{beyond-efficiency}. Explore adaptive parameter-efficient fine-tuning methods that balance security, energy efficiency, and performance on edge devices \citep{he2024deferred}.
\end{itemize}

\subsection{Hardware-Software Co-Design}
Closer integration between hardware and software development is crucial for optimizing on-device LLM performance. Future research directions include:

\begin{itemize}
    \item Advancing PIM/PNM architectures for various memory types, including optimizations for CXL-based systems and low-power solutions for edge devices \citep{kim2024breakthrough}.
    \item Developing hardware-aware optimization techniques, such as pruning-aware quantization, contextual sparsity exploitation \citep{wan2024software}, and dynamic sparse attention optimization \citep{kachris2024survey}.
    \item Enhancing AI-specific compilers and runtime systems to automatically identify and optimize operations for PIM/PNM hardware \citep{huang2024new}, considering both graph-level and hardware-specific optimizations \citep{kim2024breakthrough, wan2024software}.
    \item Designing efficient strategies for edge computing and multi-device systems, including dynamic sparse tree optimization \citep{luk2024hardware}, adaptive bit-width techniques, and energy-aware co-design approaches.
\end{itemize}

\subsection{Robustness and Reliability}
Ensuring the robustness and reliability of on-device language models under various operating conditions is paramount for their widespread adoption. Future work should address:

\begin{itemize}
    \item Investigating methods for detecting and mitigating potential biases and hallucinations in on-device LLM outputs, particularly in safety-critical applications \citep{ailem2024examining}.
    \item Exploring formal verification and validation frameworks for assessing the reliability of on-device language models in real-world scenarios \citep{remarkllm}.
    \item Leveraging ensemble methods for variance and bias reduction \citep{ensemble, ensemble1}. Exploring probabilistic inference methods to quantify and propagate uncertainty through the LLM pipeline. 

\end{itemize}

\subsection{Scalability and Deployment Optimization}
Efficiently scaling on-device LLMs to support a growing number of users and applications presents significant challenges. Future research should explore:

\begin{itemize}
    \item Developing dynamic resource allocation and load balancing techniques for distributed LLM inference across heterogeneous edge devices \citep{perllm, wilkins2024offline}.
    \item Investigating optimization strategies for reducing latency and improving throughput in collaborative edge computing scenarios, potentially leveraging techniques such as model sharding and pipelined inference \citep{EdgeShard, dhar2024empirical}. 
    \item Exploring efficient methods for managing and updating multiple LLM versions across diverse edge devices, considering factors such as network constraints and device capabilities. Building cyber-infrastructure to enhance the reusibility and reproducibility of models and datasets \citep{huggingface, datasets, core}.
\end{itemize}

\subsection{Continual Learning and Personalization}
The deployment of on-device LLMs offers unprecedented opportunities for personalized AI experiences. However, it also presents unique challenges in maintaining model relevance and adapting to new information and user preferences over time. Future research should focus on:

\begin{itemize}
    \item Implementing controllable knowledge retention and forgetting, such as selectively retaining or forgetting information as the model encounters new data streams. This is crucial for managing misinformation and ensuring ongoing accuracy. Enhance the model's ability to autonomously learn new skills and improve existing capabilities based on user interactions and local data \citep{li2024personal}. Develop effective history-tracking mechanisms to understand the evolution of the LLM through various learning phases \citep{qi2024interactive}.
    \item Advancing theoretical foundations and practical optimizations by developing robust theoretical foundations for understanding and predicting the behavior of continually learning LLMs in on-device settings. This also includes conducting large-scale user studies to refine personalization frameworks and determine effective service delivery across diverse user groups and scenarios \citep{zhang2024enabling}, as well as improving key generation and retrieval processes for better representation of task distributions in the vector space \citep{peng2024scalable}.
    \item Developing efficient continual learning mechanisms, including sophisticated data mixing strategies and efficient replay sample selection \citep{shi2024continual}. This includes exploring controllable memory systems and designing adaptive fine-tuning mechanisms for continuous model adaptation \citep{wu2024continual, li2024personal}.
\end{itemize}

Looking ahead at these future pathways and unresolved issues \citep{gao2024llm, llm-forecast, schwartz2023enhancing, mahmood2023llm, zhao2024opening}, researchers and practitioners have the opportunity to propel the on-device LLMs to new heights and transform the landscape of edge computing. The effective progression and integration of these technologies hold the potential to unlock innovative frameworks for intelligent and tailored applications, all the while tackling crucial issues surrounding security, efficiency, and dependability. The impact of these advancements reaches well beyond theoretical enhancements, offering the potential for substantial transformation across a broad spectrum of fields.
In the realm of mobile computing, enhanced on-device LLM-based AI agents \citep{chen2024octopus4} have the potential to facilitate advanced natural language interfaces and context-aware services, thereby significantly enhancing user experiences. In the context of IoT applications, these advancements empower more autonomous and adaptable systems capable of processing complex linguistic inputs in real time, even within resource-constrained environments. Within the automotive sector, improved on-device LLMs could elevate human-machine interactions in autonomous vehicles. Moreover, these technologies could enable more personalized and responsive AI-assisted patient care in healthcare.

These advancements are realized to democratize access to sophisticated AI capabilities, making them more accessible and efficient across a wide range of devices and use cases. Therefore, continued research and development in this field is both technologically imperative and socially significant, promising to herald a new era of more accessible, efficient, and reliable AI-powered applications poised to impact various facets of society and industry positively.

\section{Conclusion}
\label{sec: con}
This comprehensive review has illuminated the state-of-the-art in on-device language models. The extensive analysis presented herein has highlighted significant advancements in model compression techniques, efficient architectural designs, and hardware-software co-optimization strategies, all of which collectively facilitate the deployment of sophisticated language models on resource-constrained edge devices. The potential impact of these improvements is extensive, equipping improved data protection, decreased delay, and equal access to advanced AI capabilities across different industries and applications.

The transition from cloud-centric to edge-based LLM deployment signifies more than a mere technological progression; it represents a shift of human-AI interaction paradigms. By bringing advanced natural language processing capabilities directly to end-user devices, this transformation opens new avenues for personalized, context-aware, and instant AI experiences. On-device LLMs will revolutionize user interactions and facilitate more intelligent, responsive technologies, from mobile phones and the IoT to healthcare and autonomous systems.

However, the trajectory towards ubiquitous on-device LLMs has significant challenges. Striking an optimal balance between model performance and the inherent resource limitations of edge devices remains a critical research problem. Ensuring model robustness across heterogeneous operating conditions and developing effective continual learning mechanisms present additional hurdles. Furthermore, as the boundaries of on-device AI are pushed, questions about energy efficiency, sustainability, and responsible deployment become increasingly salient, necessitating innovative solutions and careful ethical considerations.

Realizing the full potential of on-device language models requires a concerted, multidisciplinary effort. The research community must continue advancing the frontiers of model compression techniques and efficient architecture design while concurrently addressing potential issues of data security and system reliability. Practitioners in the field should explore novel hardware-software co-design methodologies and adaptive edge-cloud collaboration strategies to optimize real-world deployments. Industry stakeholders play a pivotal role in developing specialized hardware accelerators and promoting open standards for on-device AI deployment. 

As research in this area evolves, on-device language models are positioned at the forefront of imminent technological breakthroughs. The convergence of increasingly efficient models, more powerful edge hardware, and innovative deployment strategies promises to unlock unprecedented possibilities in human-AI interaction. By addressing the challenges and capitalizing on the opportunities in this survey, the research community can work towards a future where sophisticated AI capabilities are seamlessly integrated into daily life, augmenting human abilities while respecting personalization and individuality. The journey towards ubiquitous, intelligent computing is well underway, and on-device LLMs are poised to play a pivotal role in shaping this exciting future.

In conclusion, this review serves as a comprehensive resource for researchers and practitioners, thoroughly analyzing the current state of on-device LLMs and illuminating critical areas for future research and development. As the field of on-device LLMs continues to evolve rapidly, it is imperative that the research community remains committed to addressing the challenges and embracing the opportunities presented by this transformative technology.

\bibliography{colm2024_conference}
\bibliographystyle{colm2024_conference}

\end{document}